\documentclass[letterpaper, 10 pt, conference]{ieeeconf}  %
\IEEEoverridecommandlockouts                              %

\usepackage[colorlinks,bookmarksopen,bookmarksnumbered,citecolor=red,urlcolor=red]{hyperref}
\hypersetup
{
	pdftitle = {Differential Dynamic Programming for Multi-Phase Rigid Contact Dynamics},
	pdfauthor = {Rohan Budhiraja, Justin Carpentier, Carlos Mastalli, Nicolas Mansard},
	pdfsubject = {Humanoids manuscript},
  pdfkeywords = {optimal control, legged robots, differential dynamic programming,
  rigid contacts},
	pdftoolbar = true,
	colorlinks = true,
	linkcolor = black,%
	citecolor = black,%
	urlcolor = black,
}
\usepackage[pdftex]{graphicx} %
\usepackage{subfig}
\usepackage{color}
\usepackage[tight]{units}
\usepackage{times} %
\usepackage{bm}
\usepackage{amsmath} %
\usepackage{amssymb}  %
\usepackage{glossaries}
  \newacronym{com}{CoM}{Center of Mass}
  \newacronym{dof}{DoF}{Degrees of Freedom}
  \newacronym{ddp}{DDP}{Differential Dynamic Programming}
  \newacronym{mpc}{MPC}{Model Predictive Control}
  \newacronym{oc}{OC}{Optimal Control}
  \newacronym{kkt}{KKT}{Karush-Kuhn-Tucker}
  \newacronym{grfs}{GRFs}{Ground Reaction Forces}
  \newacronym{ik}{IK}{Inverse Kinematics}
  \newacronym{id}{ID}{Inverse Dynamics}
  \newacronym{am}{AM}{Angular Momentum}

\usepackage{flushend}

\title{\LARGE \bf
Differential Dynamic Programming\\ for Multi-Phase Rigid Contact Dynamics
}

\author{Rohan Budhiraja\,$^1$, Justin Carpentier\,$^{1,2,3}$, Carlos Mastalli\,$^1$
and Nicolas Mansard\,$^1$%
\thanks{$^1$~CNRS, LAAS, 7 Avenue du Colonel Roche, Toulouse, France.}
\thanks{$^2$~D\'epartement d'informatique de l'ENS, \'Ecole Normale Sup\'erieure, CNRS, PSL Research University, 
Paris, France.}
\thanks{$^3$~INRIA, France. }
\thanks{\textit{email}: \href{mailto:rohan.budhiraja@laas.fr}{rohan.budhiraja}@laas.fr.}
}
\DeclareMathOperator*{\argmin}{arg\,min}

\begin{document}

\maketitle
\thispagestyle{empty}
\pagestyle{empty}

\newcommand{\ddqnu}{\dot{\mathbf{\nu}}}
\newcommand{\dqnu}{\mathbf{\nu}}
\newcommand{\miner}{\mathbf{M}}
\newcommand{\nonlt}{\mathbf{b}}
\newcommand{\selm}{\mathbf{S}}
\newcommand{\torq}{\boldsymbol{\tau}}
\newcommand{\forc}{\boldsymbol{\lambda}}
\newcommand{\jacco}{\mathbf{J}^{\top}_c}
\newcommand{\jaca}{\mathbf{J}}
\newcommand{\djaca}{\dot{\mathbf{J}}}
\newcommand{\zerosm}{\mathbf{0}}
\newcommand{\state}{\mathbf{x}}
\newcommand{\x}{\mathbf{x}}
\newcommand{\ctrl}{\mathbf{u}}
\newcommand{\ctrls}{\mathbf{U}}
\newcommand{\dynsys}{\mathbf{f}}
\newcommand{\f}{\mathbf{f}}
\newcommand{\dynforce}{\mathbf{g}}
\newcommand{\g}{\mathbf{g}}
\newcommand{\costf}{J}
\newcommand{\delstate}{\delta\state}
\newcommand{\delctrl}{\delta\ctrl}
\newcommand{\mq}{\mathbf{q}}
\newcommand{\mdq}{\mathbf{v}}
\newcommand{\dq}{\dot{\mathbf{q}}}
\newcommand{\mddq}{\dot{\mathbf{v}}}
\newcommand{\acc}{\mathbf{a}}
\newcommand{\holconst}{\boldsymbol{\phi}(\mathbf{q})}
\newcommand{\accfree}{\mathbf{a}_{\text{free}}}
\newcommand{\norm}[1]{{\left\lVert#1\right\rVert}^{2}}
\newcommand{\R}[1]{\mathbb{R}^{#1}}
\newcommand{\Z}[1]{\mathbb{Z}^{#1}}
\newcommand{\Q}{\mathbf{Q}}
\newcommand{\vc}[1]{\mathbf{\mathbf{#1}}}
\newcommand{\mat}[1]{\ensuremath{\begin{bmatrix}#1\end{bmatrix}}}
\newcommand{\crossmx}[1]{\mat{#1}_{\times}}

\newcommand{\sref}[1]{Section~\ref{#1}}
\newcommand{\eref}[1]{(\ref{#1})}
\newcommand{\fref}[1]{Fig.~\ref{#1}}
\newcommand{\tref}[1]{Table~\ref{#1}}

\newcommand{\alert}[1]{\textcolor{red}{\textbf{#1}}}

\begin{abstract}
A common strategy to generate efficient locomotion movements is to split the
problem into two consecutive steps: the first one generates the contact sequence together with the centroidal
trajectory, while the second step computes the whole-body trajectory that follows the
centroidal pattern. While the second step is generally handled by a simple program
such as an inverse kinematics solver, we propose in this paper to compute the
whole-body trajectory by using a local optimal control solver, namely \acrfull{ddp}.
Our method produces more efficient motions, with lower forces and smaller
impacts, by exploiting the \gls{am}. With this aim, we propose an original
\acrshort{ddp} formulation exploiting the \acrlong{kkt} constraint of the
rigid contact model. We experimentally show
the importance of this approach by executing large steps walking on the real HRP-2 robot,
and by solving the problem of attitude control under the absence of external contact forces.
\end{abstract}

\section{Introduction}

\subsection{Goal of the paper}

Trajectory optimization based on reduced centroidal dynamics~\cite{Orin2013} has gained
a lot of attention in the legged robotics community. Some approaches use it after
precomputing the contact sequence and placements
\cite{Dai2014,Carpentier2016,Carpentier2017a,Fernbach2018,Herzog2015,Winkler2015} while other strategies
optimize the centroidal trajectory and contact information together
\cite{Mastalli2017a,Winkler2018,Aceituno2017}.
In both cases, the transfer from centroidal dynamics to whole-body dynamics is achieved
using instantaneous feedback linearization to locally take into account the constraints of the robot.
These solvers usually solve quadratic optimization problems written with task-space dynamics (\gls{ik} / \gls{id})~\cite{Saab2013,Herzog2016a, Vaillant2016}.
While this scheme has shown great experimental results
(e.g.~\cite{Carpentier2016, Fernbach2018,Mastalli2017b}), it is still not able to correctly
handle the \gls{am} produced by the extremities of the limbs.
This is notable for humanoid robots which have important masses in the limbs that are put in motion during (for instance) walking.
This effect is neither properly captured by the centroidal model,
nor by the instantaneous time-invariant linearization.

In
\cite{Herzog2016a} an alternative scheme aims to compensate the \gls{am}
variations. Indeed, it properly compensates the momentum changes produced by the flying
limbs, however it is not yet able to trigger additional momentum to enable very
dynamic movements. This would be needed for generating long steps,
running, jumping or salto motions.
To properly handle the \gls{am}, it is necessary to jointly optimize the
whole-body kinematics and the centroidal dynamics~\cite{Dai2014}. However
whole-body trajectory optimization approaches suffer from two problems that prevent
the replacement of \gls{ik}/\gls{id} solvers. Namely, they struggle to discover a valid
motion, in particular the gait and its
timings; and they are slow to converge.

In this paper, we propose to combine the advantages of centroidal dynamics
optimization (to decide the gait, the timings and the main shape of the
centroidal trajectory) with a whole-body trajectory optimizer based on multi-phase
rigid contact dynamics. 
In what follows, we first discuss the importance of properly handling
the \gls{am} during locomotion, before introducing our method.

\subsection{On the importance of angular momentum}

Consider an astronaut, floating in space, without any external forces. If
he/she mimics the normal human walk, he/she will start spinning in his/her sagittal plane.
Indeed, contact forces are
not the only way to change the robot orientation.
It is known \cite{Wieber2006} that robot orientation can be controlled without the need of contact forces
(i.e. only with the internal joint actuators).
Under the action of only internal forces, the \gls{am}
conservation can be seen as a non-holonomic constraint on the robot orientation.
Of course, one can design a control law that counterbalances the lower-body
\gls{am}. However this will create tracking errors
(and potentially instabilities) without mentioning the cases where the arms
need to be used for multi-contact locomotion.
In fact, as shown in~\cite{Brockett1983}, a system under non-holonomic constraints
cannot be controlled with a time-invariant feedback law.
Thus \gls{am} requires a
preview control strategy to be correctly regulated or triggered.

It is often (wrongly) understood that centroidal optimization provides the answer to this
problem. The centroidal optimizer can neither anticipate nor modify the limb movements 
in order to change the \gls{am} as needed. For instance, the centroidal
optimizer cannot anticipate a high
demand of the linear part (\gls{com}) by delaying the limb movement, or exploit the
movement of the arms to compensate for large forces acting for a short
duration. Nonetheless, these methods are still valid since they provide an efficient
way to compute the \gls{com} motion while keeping balance and avoiding slippage.

\begin{figure*}%
\centering
\includegraphics[width=0.98\textwidth]{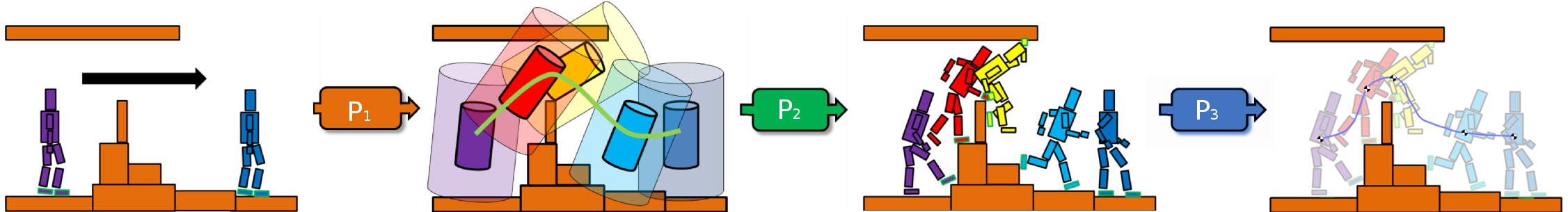}
\caption{Overview of our multi-stage locomotion framework \cite{Carpentier2017}. 
Given a requested path request between start and goal positions (left image),
$\mathcal{P}_1$ is the problem of computing a guide path in the space of
equilibrium feasible root configurations. $\mathcal{P}_2$ is then the problem of extending the path
into a discrete sequence of contact configurations. Finally, $\mathcal{P}_3$ attends
to compute a dynamic-physical whole-body trajectory given as input the discrete contact
sequence.}
\label{fig:loco3d_pipeline}
\end{figure*}

\subsection{Overview of our method}

Instead of relying on instant linearization using \gls{ik}/\gls{id}, 
we propose to rely on optimal control~\cite{Lengagne2013}~\cite{Schultz2010},
namely \gls{ddp}, to
compute the whole-body motion while tracking the centroidal trajectory.
\gls{ddp} has been
made popular by the proof of concept \cite{Koenemann2015}, and by the demonstration
in simulation that it can meet the control-loop timings constraint \cite{Tassa2014b}.
However, locomotion movements computed by \gls{ddp} have not yet been transferred
to a real full-size humanoid. Contrary to \cite{Koenemann2015} that optimizes
the motion from scratch with a regularized dynamics (thanks to a  smooth contact model
\cite{MuJoCo}), we propose to impose the contact phases as decided by
the centroidal optimization. As \gls{ddp} does not need to
discover the contact switching instants, we can use rigid contact
dynamics which is faster to compute and easier to implement.

Other works have shown that \gls{ddp} is able to discover locomotion
gaits applied on a real quadruped ~\cite{Neunert2018}. In~\cite{Rajamaki2016},
\gls{ddp} is coupled with Monte Carlo tree search to compute the bipedal locomotion pattern of an avatar.
While not yet demonstrated on a real humanoid, we might wonder whether this
should be pushed further, instead of relying on a decoupling between contact
computation, centroidal and whole-body optimization. We believe that \gls{ddp}
is a mature solution to replace \gls{ik}/\gls{id} and is very complementary to centroidal
optimization. Indeed, contact and centroidal problems can be efficiently
handled within a global search thanks to the low dimension, while \gls{ddp}
is efficient to accurately handle the whole-body dynamics in a large space
(but locally).

The rest of the paper is organized as follows: after discussing the
locomotion framework in which our method takes place, we describe and
justify our technical choices in \sref{sec:locomotion}. \sref{sec:ddp}
briefly introduces the \gls{ddp} algorithm, we then
describe our novel \gls{ddp} formulation for rigid contact dynamics
in \sref{sec:constrained_ddp}. Then, in \sref{sec:results} we show 
experimental trials and realistic simulation on the HRP-2 robot and
compare them against a whole-body \gls{ik} solver. Lastly, \sref{sec:conclusions}
summarizes the work conclusions.

\section{Multi-contact Motion Generation}\label{sec:locomotion}

Locomotion synthesis is a difficult problem because of a) the combinatorial nature of
contact planning, b) the high-dimensionality of the search-space, c) the
instabilities, discontinuities and non-convexity of the robot dynamics,
d) the non-convexity of the terrain environment, among others. 
In our previous works, we have proposed a multi-stage strategy that decouples the global
problem into successive subproblems of smaller dimensions~\cite{Carpentier2017}.
The global problem is thus split into an interactive acyclic contact
planner~\cite{Tonneau2018}, a centroidal pattern generator~\cite{Carpentier2017a}~\cite{Fernbach2018}
which takes the contact sequence as an input, and a whole-body motion generator.

Centroidal pattern generator~\cite{Carpentier2017a}~\cite{Fernbach2018} by itself is unable to account for
the \gls{am} effect generated by the limb motions. Indeed, \gls{am} of a body is accounted by both, as a result of
the contact forces, and by \emph{only} the limb movement. Consider a floating-base robot.
Even with no external forces acting on the robot,
a constant \gls{am} can be maintained by the non-holonomic constraint on the joint velocities~\cite{Wieber2006}:

\begin{equation}
\sum_{k=0}^{n_j} m_i[\vc{r}_k-\vc{r}]_{\times}\dot{\vc{x}}_k +
 \vc{R}_k\vc{I}_k\boldsymbol{\omega}_k=\textrm{Constant},
\label{eq:angular_momentum_conservation}
\end{equation}
where $k$ denotes the index of a rigid limb and $\vc{I}_k$ corresponds to its inertia matrix
expressed in the body's \gls{com} frame. $\dot{\vc{x}}_k$ and $\boldsymbol{\omega}_k$ are
the linear and angular velocities of the links.

While we would like to account for this ``gesticulation''~\cite{Wieber2006} during the centroidal optimization,
it is a difficult problem to solve in real-time. Instead, we assume that this
effect is small~\cite{Carpentier2017a}, but
yet important to consider, and we track it with a whole-body
\gls{mpc}. \gls{ddp}~\cite{Mayne1973} is a reasonable choice between
the two, because it allows us to
generate additional \gls{am} using \eref{eq:angular_momentum_conservation}, while efficiently
tracking the reference trajectories provided by the centroidal solver.
\gls{ddp} has been shown~\cite{Tassa2012} to be efficient in solving online \gls{oc} in legged systems.
\fref{fig:loco3d_pipeline} demonstrates our multi-stage locomotion
pipeline.

Generating a whole-body trajectory requires finding a trajectory
which is subject to dynamic-consistency, the friction-cone
constraints, the self-collision avoidance and the joint limits. This can be formulated as a
single \gls{oc} problem:
\begin{equation}
\label{eq:oc_problem}
\begin{aligned}
\begin{Bmatrix} \state^*_0,\cdots,\state^*_N \\
				  \ctrl^*_0,\cdots,\ctrl^*_N
\end{Bmatrix} =
\argmin_{\vc{X},\vc{U}}
&
& & \sum_{k=1}^N \int_{t_k}^{t_k+\Delta t} l_k(\vc{x},\vc{u})dt \\
& \text{s.t.}
& &   \dot{\state} = \dynsys(\state,\ctrl),\\
& & & \state\in\mathcal{X}, \ctrl\in\mathcal{U}, \forc\in\mathcal{K}.
\end{aligned}
\end{equation}

The state and control are defined as \mbox{$\vc{x}=(\mq,\mdq)$} and
$\vc{u}=\torq$, where $\mq\in SE(3)\times \R{n_j}$ is the configuration vector for a floating base robot with $n_j$ \gls{dof}.
$\mdq$ is its derivative, $\torq$ is the vector of joint torques,
and $\forc$ is the friction force corresponding to $(\vc{x},\vc{u})$.
$\mathcal{X}$, $\mathcal{U}$ and $\mathcal{K}$ are the admissible sets: joint configurations and joint
velocities bounds, joint torque commands limits, and contact forces constraints (e.g. the friction
cone constraint), respectively.

\section{Differential Dynamic Programming}\label{sec:ddp}
In this section, we give a formal description of the \gls{ddp} algorithm for 
completeness. For more elaborate explanations and derivations, the reader is
referred to \cite{Mayne1973}. \gls{ddp} belongs to the family of \gls{oc} handled with a sparse structure thanks to the Bellman principle. Concretely, instead of finding the entire optimal trajectory \eref{eq:oc_problem},
we make recursive decisions:
\begin{equation}
  \begin{split}
    V_i(\state_i) = \min_{\ctrl_i}\,[l(\state_i,\ctrl_i)+V_{i+1}(\dynsys(\state_i,\ctrl_i))],
  \end{split}
  \label{eq:valuef_rec}
\end{equation}
This is possible through a forward simulation of the system dynamics
$\state_{i+1} = \dynsys(\state_i,\ctrl_i)$. Note that $V_i$ denotes the value
function which describes the minimum cost-to-go:
\begin{equation}
  \begin{split}
    V_i(\state_i) = \min_{\ctrl_{i:N-1}} \costf_i(\state_{i},\ctrl_{i:N-1}).
  \end{split}
  \label{eq:valuef}
\end{equation}
\gls{ddp} searches locally the optimal state and control sequences of the above
problem. It uses a quadratic approximation $\vc{Q}(\delta\state,\delta\ctrl)$
of the differential change in \eref{eq:valuef_rec}, i.e.
\begin{equation}
  \vc{Q}(\delta\state,\delta\ctrl) \approx
  \mat{1 \\ \delta\state \\ \delta\ctrl}^\top
  \mat{\vc{0}  &  \vc{Q_x}^\top  &  \vc{Q_u}^\top \\
       \vc{Q_x}  &  \vc{Q_{xx}}  &  \vc{Q_{xu}} \\
       \vc{Q_u}  &  \vc{Q_{ux}}  &  \vc{Q_{uu}} }
  \mat{1 \\ \delta\state \\ \delta\ctrl}
\end{equation}
where
\begin{equation}
  \begin{split}
    \vc{Q_{x}} &= \vc{l_{x}} + \vc{f}^{\top}_\vc{x}\vc{V_{x}'},\\
    \vc{Q_{u}} &= \vc{l_{u}} + \vc{f}^{\top}_\vc{u}\vc{V_{x}'},\\
    \vc{Q_{xx}} &= \vc{l_{xx}} + \vc{f}^{\top}_\vc{x}\vc{V_{xx}'}\vc{f_x} + \vc{V}'_{\vc{x}}\vc{f_{xx}},\\
    \vc{Q_{uu}} &= \vc{l_{uu}} + \vc{f}^{\top}_\vc{u}\vc{V_{xx}'}\vc{f_u} + \vc{V}'_{\vc{x}}\vc{f_{uu}},\\
    \vc{Q_{ux}} &= \vc{l_{ux}} + \vc{f}^{\top}_\vc{u}\vc{V_{xx}'}\vc{f_x} + \vc{V}'_{\vc{x}}\vc{f_{ux}},
  \end{split}
  \label{eq:hessian_q}
\end{equation}
and the primes denotes the values at the next time-step.

\subsection{Backward pass}
The backward pass determines the search direction of the Newton step by recursively solving \eref{eq:valuef_rec}. In an unconstrained setting the solution is:
\begin{equation}
  \delctrl^* = \argmin_{\delctrl} \vc{Q}(\delstate,\delctrl) = \vc{k} + \vc{K}\delstate,\\
  \label{eq:compute_deltau}
\end{equation}
where $\vc{k}=-\vc{Q}^{-1}_{\vc{uu}}\vc{Q_{u}}$ and
$\vc{K}=-\vc{Q}^{-1}_{\vc{uu}}\vc{Q_{ux}}$ are the feed-forward and feedback terms.
Recursive updates of the derivatives of the value function are done as follows:
\begin{equation}
  \begin{split}
    \vc{V_{x}}(i) &= \vc{Q_{x}} + \vc{K}^{\top}\vc{Q_{uu}}\vc{k} + \vc{K}^{\top} \vc{Q_{u}} + \vc{Q}^{\top}_{\vc{ux}}\vc{k}, \\
    \vc{V_{xx}}(i) &= \vc{Q_{xx}} + \vc{K}^{\top}\vc{Q_{uu}}\vc{K} + \vc{K}^{\top} \vc{Q_{ux}} + \vc{Q}^{\top}_{\vc{ux}} \vc{K}.
  \end{split}
  \label{eq:recursive_hessian_gradient}
\end{equation}

\subsection{Forward pass}
The forward pass determines the step size along the Newton direction by adjusting
the line search parameter $\alpha$. It computes a new trajectory by integrating the
dynamics along the computed feed-forward and feedback commands $\{\vc{k}_i,\vc{K}_i\}$:
\begin{equation}
  \begin{split}
    \hat{\ctrl}_i &= \ctrl_{i} + \alpha\vc{k}_i + \vc{K}_i (\hat{\state}_i - \state_i), \\
    \hat{\state}_{i+1} &= \dynsys(\hat{\state}_i,\hat{\ctrl}_i),\\
  \end{split}
\end{equation}
in which $\hat{\state}_1 = \state_1$, and $\{\hat{\state}_i,\hat{\ctrl}_i\}$ are the new
state-control pair. Note that if $\alpha = 0$, it does not change the state and control
trajectories.

\subsection{Line search and regularization}
We perform a \textit{backtracking} line search by trying the full step
$(\alpha=1)$ first. The choice of $\alpha$ is dual to the choice of regularization
terms, and both are updated between subsequent iterations to ensure a good progress toward the (local) optimal solution. We use two regularization schemes: the Tikhonov regularization (over $\vc{Q_{uu}}$) and its update using
the Levenberg-Marquardt algorithm are typically used \cite{toussaint2017}. Tassa et al.~\cite{Tassa2012} propose a regularization scheme over $\vc{V_{xx}}$, which is
equivalent to adding a penalty in the state changes.

\section{DDP with Constrained Robot Dynamics}\label{sec:constrained_ddp}

\subsection{Contact dynamics}
Let's consider the case of rigid contact dynamics with the environment. Given a
predefined contact sequence, rigid contacts can be formulated as holonomic
scleronomic constraints to the robot dynamics (i.e. equality-constrained
dynamics). The unconstrained robot dynamics is typically represented as:
\begin{equation}
  \miner \mddq_{free} = \torq_b,
  \label{eq:dyn_no_force}
\end{equation}
where $\miner\in\R{n\times n}$ is the joint-space inertia matrix, 
$\mddq_{free}$ is the unconstrained joint acceleration vector,
\mbox{$\torq_b = \selm\torq-\nonlt\in\R{n}$} is the force-bias vector that accounts for
the control $\torq$, the Coriolis and gravitational effects $\nonlt$, and $\selm$ is the
selection matrix of the actuated joint coordinates.

We can account for the rigid contact constraints by applying the Gauss
principle of least constraint~\cite{Udwadia1992,Wieber2006}. Under
this principle, the constrained motion evolves in such a way that it
minimizes the deviation in acceleration from the unconstrained motion $\accfree$, i.e.:
\begin{equation}
  \label{eq:gauss_min}
\begin{aligned}
  & \mddq = \underset{\acc}{\argmin}
  & & \frac{1}{2}\,\|\mddq-\mddq_{free}\|_{\vc{M}} \\
  & \textrm{subject to}
  & &   \mathbf{J}_{c} \mddq + \djaca_c \mdq = \zerosm,
\end{aligned}
\end{equation}
in which $\miner$ is formally the metric tensor over the configuration
manifold $\mq$. Note that we differentiate twice the holonomic contact constraint
$\holconst$ in order to express it in the acceleration space. In other words,
the rigid contact condition is expressed by the second-order kinematic
constraints on the contact surface position. $\vc{J}_{c} =
\begin{bmatrix} \vc{J}_{c_1} & \cdots & \vc{J}_{c_f}\end{bmatrix}\in\R{kp\times n}$
is a stack of the $f$ contact Jacobians.

\subsection{\gls{kkt} conditions}
The Gauss minimization in \eref{eq:gauss_min} corresponds to an equality-constrained convex
optimization problem\footnote{$\miner$ is a positive-definite matrix.},
and it has a unique solution if $\vc{J}_{c}$ is full-rank. The primal and
dual optimal solutions $(\mddq,\forc)$ must satisfy the so-called \gls{kkt} conditions
given by
\begin{equation}
  \left[
    \begin{matrix}
      \miner & \jacco \\
      {\mathbf{J}_{c}} & \zerosm \\
    \end{matrix}
    \right]
  \left[ \begin{matrix} \mddq \\ -\forc \end{matrix} \right]
  = 
  \left[
    \begin{matrix}
      \torq_b \\
      -\djaca_c \mdq \\
    \end{matrix}
  \right].
  \label{eq:kkt_contact}
\end{equation}

These dual variables $\forc^k\in\R{p}$ render themselves nicely in
mechanics as the external forces at the contact level. 
This relationship allows us to express the contact forces directly in terms of the
robot state and actuation. As compared to previous approaches which would introduce
the contact constraints in the whole-body optimization~\cite{Carpentier2016}~\cite{Saab2013}, here we solve for the contact
constraints at the level of the dynamics, and not the solver.
In other words, this would free the solver to find an unconstrained
solution to the \gls{kkt} dynamics \eref{eq:kkt_contact}, without worrying about
the contact constraint. Fast iterative Newton and quasi-Newton methods can
then be easily applied to achieve real-time performance.

\subsection{\gls{kkt}-based \gls{ddp} algorithm}
From \eref{eq:kkt_contact}, we can see the augmented \gls{kkt} dynamics as a
function of the state $\state_i$ and the
control $\ctrl_i$:
\begin{equation}
  \begin{split}
    \state_{i+1} = \dynsys(\state_i,\ctrl_i),\\
    \forc_{i} = \dynforce(\state_i,\ctrl_i),
  \end{split}
  \label{eq:updated_state_fn}
\end{equation}
where the state $\state=(\mq, \mdq)$ is represented by the configuration vector and
its tangent velocity, $\ctrl$ is the torque-input vector, and $\dynforce(\cdot)$
is the dual solution of \eref{eq:kkt_contact}. In case of legged robots,
the placement of the free-floating link is described using the special Euclidean group $SE(3)$.

Given a reference trajectory for the contact forces, the \gls{ddp} backward-pass cost
and its respective Hessians (see \eref{eq:valuef_rec} and \eref{eq:hessian_q})
are updated as follows:
\begin{equation}
  \costf_i(\state_i,\ctrls_i) = l_f(\state_{N})+\sum_{k=i}^{N-1}l(\state_k,\ctrl_k,\forc_k),
  \label{eq:full_cost_force}
\end{equation}
where $\ctrls_i = \{\ctrl_i, \ctrl_{i+1}, \cdots, \ctrl_{N-1}\}$ is the
tuple of controls that acts on the system dynamics at time $i$, and the
Gauss-Newton approximation of the $\Q$ coefficients (i.e. first-order
approximation of $\dynforce(\cdot)$ and $\dynsys(\cdot)$) are
\begin{equation}
  \begin{split}
    \Q_{\x} &= \vc{l}_{\x} + \g_{\x}^{\top} \vc{l}_{\forc} + \f_{\x}^{\top}\vc{V_{\x}'}, \\
    \Q_{\ctrl} &= \vc{l}_{\ctrl} + \g_{\ctrl}^{\top} \vc{l}_{\forc} + \f_{\ctrl}^{\top}\vc{V_{\x}'},\\
    \Q_{\x\x} &\approx \vc{l}_{\x\x} + \g_{\x}^{\top} \vc{l}_{\forc\forc}\g_{\x} + \f_{\x}^{\top}\vc{V_{\x\x}'}\f_{\x},\\
    \Q_{\ctrl\ctrl} &\approx \vc{l}_{\ctrl\ctrl} + \g_{\ctrl}^{\top} \vc{l}_{\forc\forc}\g_{\ctrl} + \f_{\ctrl}^{\top}\vc{V_{\x\x}'}\f_{\ctrl},\\
    \Q_{\ctrl\x} &\approx \vc{l}_{\ctrl\x} + \g_{\ctrl}^{\top} \vc{l}_{\forc\forc}\g_{\x} + \f_{\ctrl}^{\top}\vc{V_{\x\x}'}\f_{\x}.\\
  \end{split}
  \label{eq:hessian_q_lambda}
\end{equation}
The set of equations (\ref{eq:hessian_q_lambda}) takes into account the
trajectory of the rigid contact forces inside the backward-pass. The system
evolution needed in the forward-pass is described by \eref{eq:updated_state_fn}.

\section{Results}\label{sec:results}

\begin{figure}%
  \centering
  \includegraphics[width=0.46\textwidth]{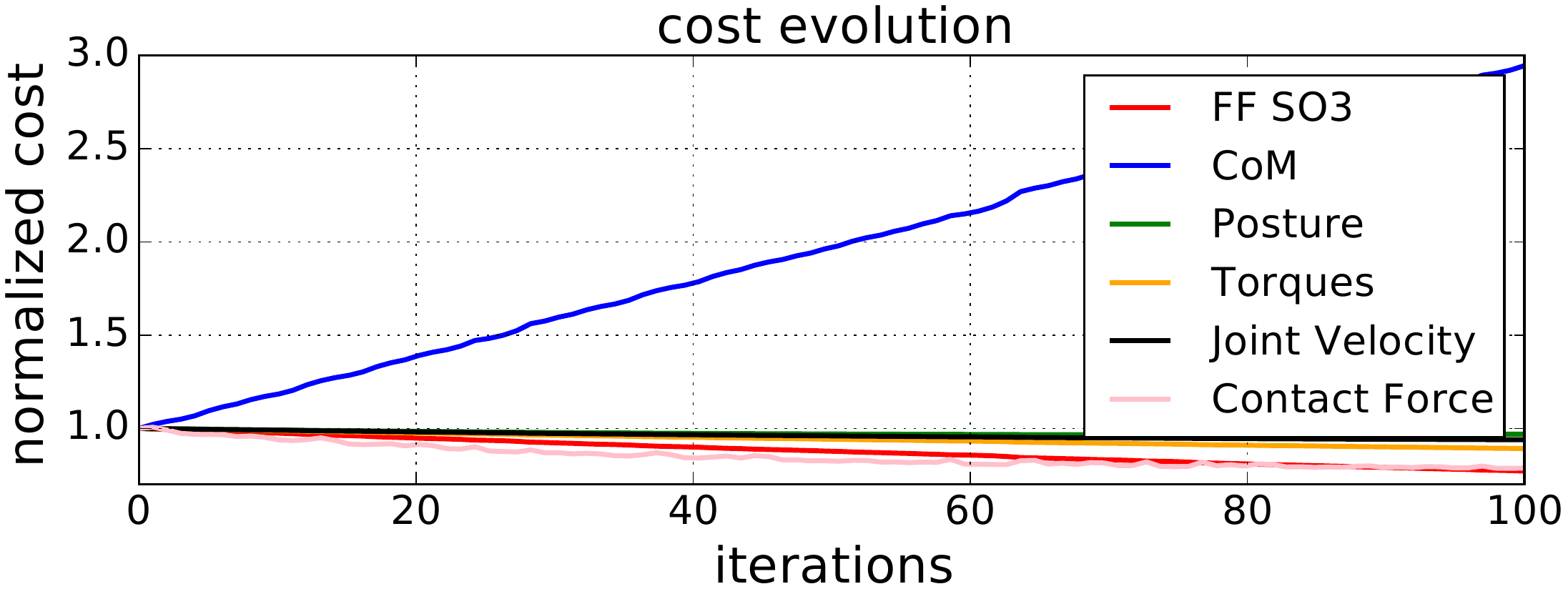}
  \caption{Evolution of the different cost functions (normalized) with respect to iterations. \gls{ddp} reduces the applied torques by recalculating the \gls{com} tracking. It improves the contact force by taking into account the whole-body angular momentum. The result is a continuous improvement in the performance as compared to \gls{ik}. We stop after 100 iterations.}
  \label{fig:cost_breakup}
  \end{figure}

\begin{figure}
\centering
\includegraphics[width=0.46\textwidth]{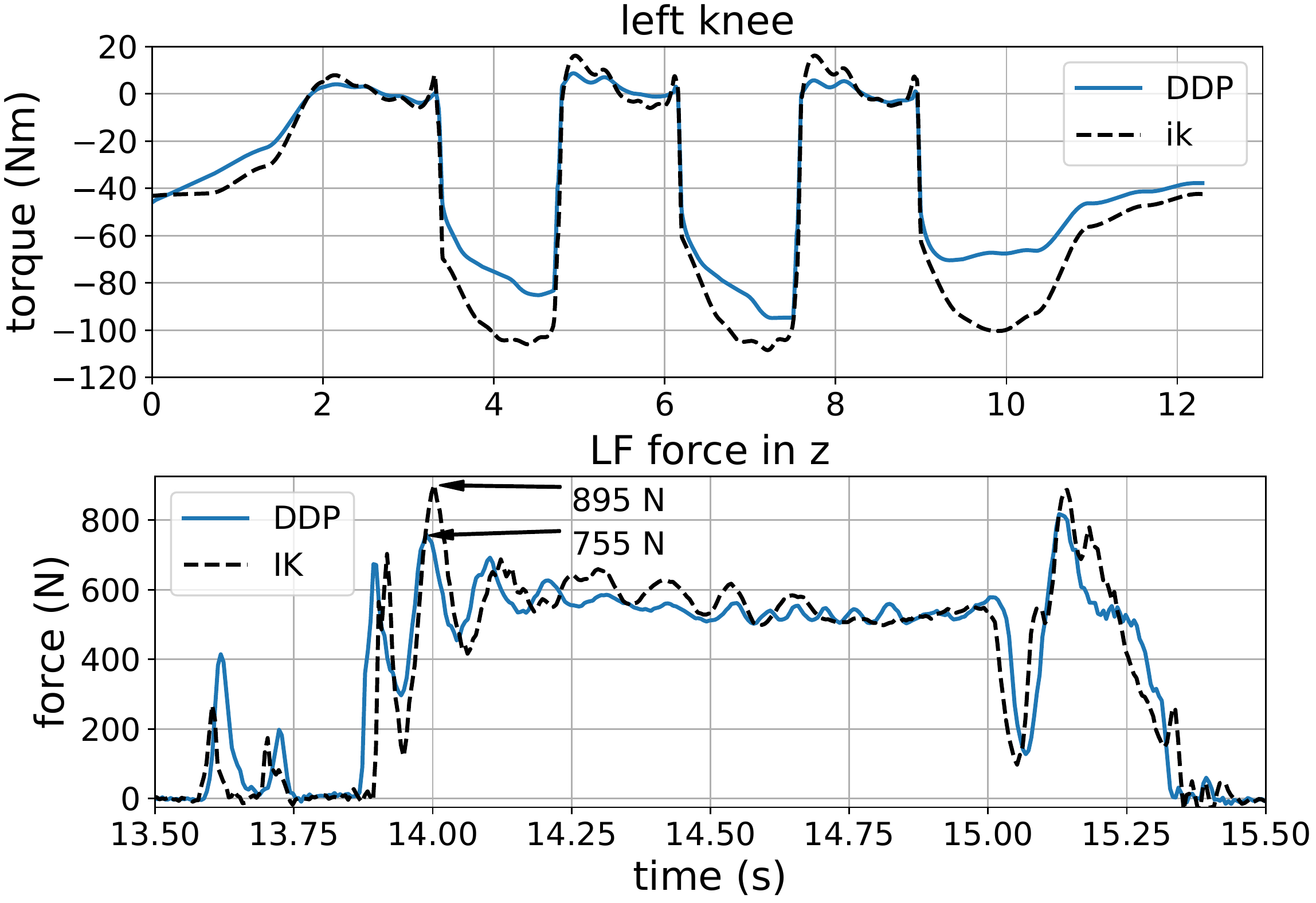}
\caption{Comparison between the \gls{ik} and \gls{ddp} trajectory for \unit[100]{cm}
  stride on the HRP-2 robot. \textit{Top}: Knee torques generated in the left leg.
  \textit{Bottom}: \gls{grfs} generated in the left foot. The \gls{ddp} formulation allows us to utilize the angular momentum of the upper body, which reduces the requirement on the lower body to create a counterbalancing motion. This results in a lower torque in the lower body, as well as lower
\gls{grfs}. Around $t=14s$ we can see high peaks for the
\gls{ik} and \gls{ddp} trajectories of \unit[895]{N} and \unit[755]{N}, respectively.}
\label{fig:grfs_comparison}
\end{figure}

\begin{figure}
  \centering
  \includegraphics[width=0.46\textwidth]{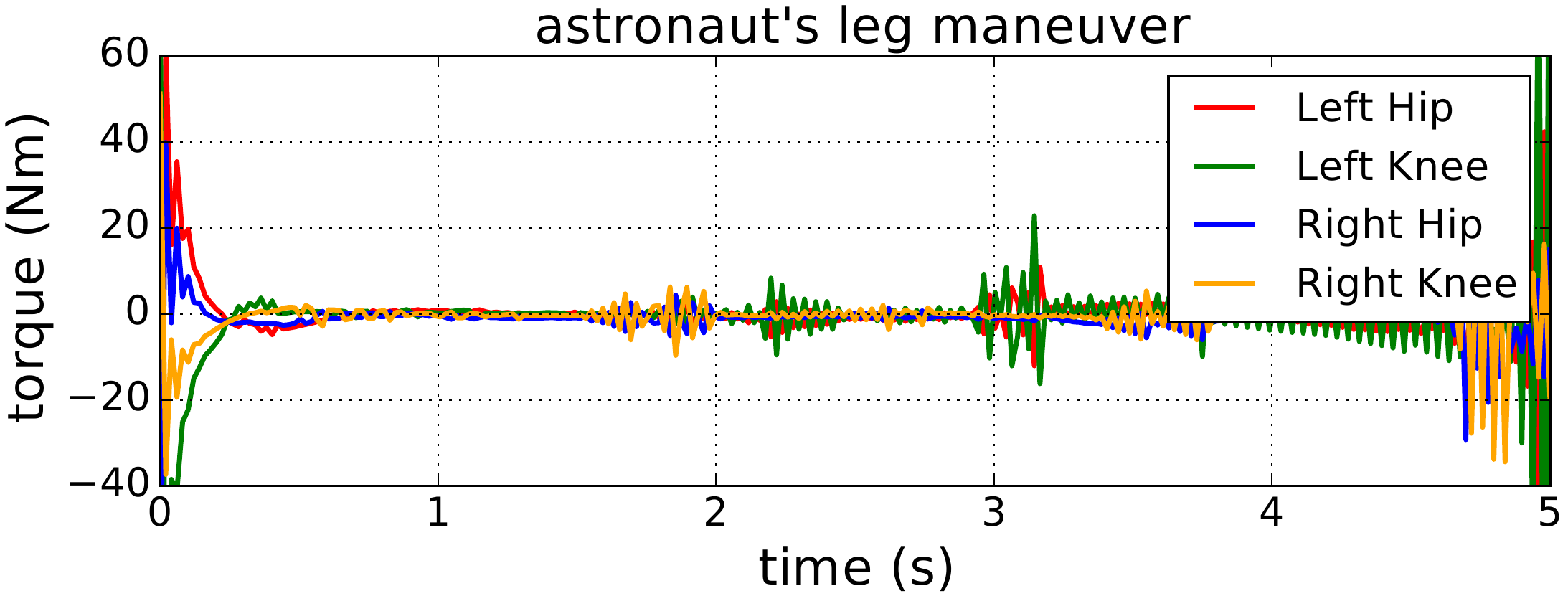}
  \caption{Joint torques for the astronaut maneuver. Our method plans a smart strategy by kick-starting the rotation, and then tries to maintain the velocity by small bang-bang control signals. Towards the end, it changes again the velocity of the lower legs in order to bring the system to a stop.}
  \label{fig:joint_torques}
\end{figure}

\begin{figure*}
\centering
\includegraphics[width=0.98\textwidth]{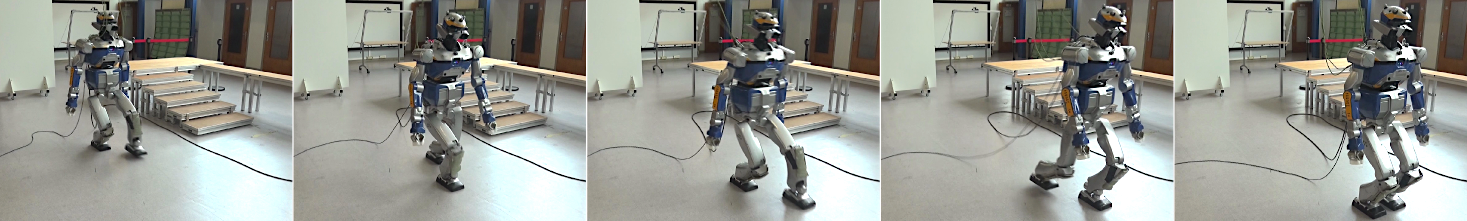}
\caption{Snapshots of \unit[100]{cm} stride on a flat terrain used to
evaluate the performance of our whole-body trajectory optimization method. The
\gls{ddp} trajectory reduces significantly the normal forces peaks compared
with classic whole-body IK.}
\label{fig:stride_walk}
\end{figure*}

\begin{figure*}
\centering
\includegraphics[width=0.98\textwidth]{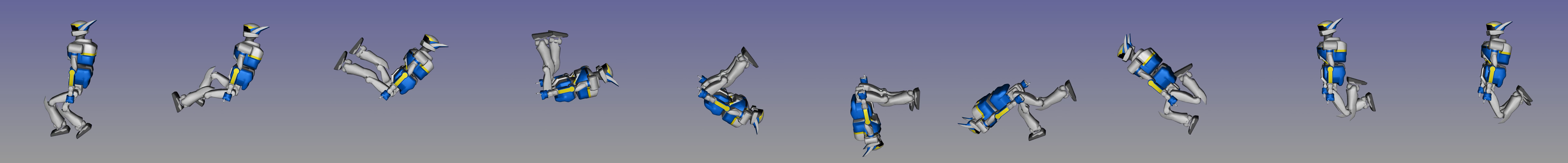}
\caption{Attitude adjustment maneuver conducted by the robot in gravity free space. \gls{ddp} solver takes into account the non-holonomic angular momentum constraint and uses internal actuation to rotate 360$^{\circ}$ without the need for contact forces.}
\label{fig:astronaut}
\end{figure*}

In this section, we show that our \gls{ddp} formulation can generate whole-body
motions which require regulation of the angular momentum.
The performance of our algorithm is assessed on realistic simulations
and aggressive experimental trials on the HRP-2 robot. First, we perform 
very large strides (from \unit[80]{} to \unit[100]{cm}) which require large amount of angular momentum (due to the fast swing of the 6-kg leg) and reach the HRP-2 limits.
This demonstrates the ability of the solver to handle contact constraints, as well as to
\textit{generate} excessive \gls{am} required by the leg motions.
Then, to again emphasize the need for a horizon based optimizer such as \gls{ddp}, and the ability of our
solver to handle these \gls{am} requirements, we regulate the robot
attitude in absence of contact forces and gravitation field.
These motions cannot be generated through a standard time-invariant \gls{ik}/\gls{id} solver,
as the system becomes non-holonomic as shown in \eref{eq:angular_momentum_conservation}.

All the motions were computed offline. Contact
sequence~\cite{Tonneau2018} and the centroidal trajectory \cite{Carpentier2018b} are precomputed and provided to the solver for the large stride experiments. We used the standard
controller OpenHRP~\cite{kanehiro2004openhrp} for tracking the motions on the real robot. The large strides produced by \gls{ddp} are compared with those produced by an \gls{ik} solver \cite{Saab2013}, showing the benefit of our approach.

\subsection{Large stride on a flat ground}
In these experiments, we generate a sequence of cyclic contact for
\unit[80]{cm} to \unit[100]{cm} stride. These are very big steps for
HRP-2 compared to its height (\unit[160]{cm}).
For the contact location, we
use the \gls{oc} solver reported in \cite{Carpentier2016} to compute the
contact timings and the centroidal trajectory. As the centroidal solver is able to provide
feasible contact forces for individual contacts in the phase, we use a damped cholesky
inverse to deal with the rank deficient $\mathbf{J}_{c}$ matrix. Then we use our proposed
\gls{ddp} to generate the full robot motion.

The cost function is composed of various quadratic residues (i.e.
$\|\vc{r}_i(\vc{x},\vc{u},\boldsymbol{\lambda})\|^2_{\vc{Q}_i}$) in order to keep balance
and to increase efficiency and stability: (a) \gls{com}, foot position and
orientation and contact forces tracking of centroidal motion, (b) torque
commands minimization and (c) joint configuration and velocity
regularization. The evolution of the different normalized task costs with iterations is shown in \fref{fig:cost_breakup}. Our method adapts the \gls{com} to create a more efficient torque and contact force trajectory. 

Increasing the upper-body angular momentum helps to counterbalance the swing leg motion, this in turn reduces \gls{grfs} and improves the locomotion stability. 
Our experimental results show a reduction on the \gls{grfs} peaks
compared to the IK solver. \fref{fig:grfs_comparison} shows the measured normal
contact forces and the knee torques in case of DDP solver and IK solvers.
Our \gls{ddp} reduced the normal forces peaks of the IK solver from 
\unit[895]{N} to \unit[755]{N}. This represents a significant improvement, considering
that the minimum possible contact forces are \unit[650]{N} (the total mass of the HRP-2
robot is \unit[65]{kg}) and the maximum safe force allowed by the sensors on the foot is \unit[1000]{N}.
An overview of the motion is shown in \fref{fig:stride_walk}.

\subsection{Attitude regulation through joint motion}

The angular momentum equation \eref{eq:angular_momentum_conservation}
shows that it is possible to regulate the robot attitude without the need of
contact forces \cite{Wieber2006}. It can be seen that the gravity field does
not affect this property. Thus, we analyze how our \gls{ddp} solver regulates
the attitude in zero-gravity condition, we named this task \textit{astronaut
reorientation}. The astronaut reorientation (similar to cat falling) is an
interesting motor task due to fact that it depends on a proper exploitation of
the angular momentum based on the coordination of arms and legs motions.
\fref{fig:astronaut} demonstrates the motion found by the solver to rotate the body 360$^{\circ}$.
Unlike an instantaneous tracking solver like IK, the solver is willing to bend in the opposite direction, in order to
obtain an ability to create sufficient angular momentum by the legs. 
It is important here to note that such motion cannot be obtained by a time-invariant control law which does not take the future
control trajectory into account.

The cost matrices for this problem require a barrier function on the robot configuration to avoid self-collision.
Final cost on the body orientation provides the goal, and a running cost on the posture is added for regularization. No warm start is given to the
solver, the initial control trajectory is a set of zero vectors.
For the ease of demonstration, we used only the leg joints in the sagittal plane.
\fref{fig:joint_torques} shows the torques produced by the hip and the knee joints.
Our method creates a rotation of the upper body by a quick initial motion in the legs.
Then it maintains the angular velocity by small correctional torque inputs during the rest of the trajectory.
At the end, to bring the rotation to a halt, the same behavior is repeated in the reverse.

\section{Conclusion and Perspective}\label{sec:conclusions}
Typically, reduced centroidal trajectory optimization does not take into account
the \gls{am} produced by the limb motions. Proper regulation of the
\gls{am} exploits the counterbalancing effect of limb motion in order to reduce the
contact forces and torque inputs. It also improves the stability during
flight phases where the momentum control can only be made through joint motions.
Excessive \gls{am} cannot be produced by a simple \gls{ik}/\gls{id} solver. \gls{oc} provides
the required tools for solving it. Our proposed solver is an extension of our previous work \cite{Carpentier2016}.

In this paper, we have proposed a novel \gls{ddp} formulation based on the
augmented \gls{kkt} dynamics (see \eref{eq:updated_state_fn}) which is a product
of holonomic contact constraints. It represents the first application of motion generated by \gls{ddp} solver on a real humanoid locomotion. Our whole-body motion generation pipeline
enables us to potentially regulate angular momentum dynamics during the whole-body motion in
real-time. We have observed a reduction of the contact forces compared to the \gls{ik}
solver, even though we had to restrict the angular momentum in the sagittal
plane for the stride on flat ground task due to robot limits. A more revealing experiment, 
the \textit{astronaut reorientation}, demonstrates further the limits of the previous approaches and the advantages of using \gls{ddp}. The solver generates the desired motion from scratch in this case by manipulating joint velocities within the non-holonomic angular momentum constraints.

While the solver is able to generate \gls{am}, and track the centroidal trajectories,
the current approach still lacks an assurance that the additional \gls{am} generated
by the \gls{ddp} solver is accounted for in the centroidal optimization, and vice versa.
Upcoming work from our team is focused on this guarantee.
While this is proof-of-concept, an implementation based on Analytical Derivatives~\cite{Carpentier2018} is also our next goal.

\addtolength{\textheight}{-13cm}   %

\bibliographystyle{IEEEtran}
\bibliography{library}

\end{document}